\documentclass[conference]{IEEEtran}
\IEEEoverridecommandlockouts
\usepackage{cite}
\usepackage{amsmath,amssymb,amsfonts}
\usepackage{multirow}
\usepackage{algorithmic}
\usepackage{graphicx}
\usepackage{textcomp}
\usepackage{float}
\usepackage{comment}

\usepackage[dvipsnames]{xcolor}
\usepackage{caption}

\usepackage[utf8x]{inputenc}
\def\BibTeX{{\rm B\kern-.05em{\sc i\kern-.025em b}\kern-.08em
    T\kern-.1667em\lower.7ex\hbox{E}\kern-.125emX}}

\begin{document}

\title{Auto-Assembly: a framework for automated robotic assembly directly from CAD.\\}


\author{\IEEEauthorblockN{
Fedor Chervinskii\textsuperscript{\textsection},
Sergei Zobov\textsuperscript{\textsection},
Aleksandr Rybnikov\textsuperscript{\textsection},
Danil Petrov\textsuperscript{\textsection},
Komal Vendidandi\textsuperscript{\textsection}
}
\IEEEauthorblockA{$\Lambda$ $\Gamma$ $\Gamma$ I V $\Lambda$ L}}

\maketitle
\begingroup\renewcommand\thefootnote{\textsection}
\footnotetext{Equal contribution. \newline [chervinskii, zobov, rybnikov, danil.petrov, vendidandi]@arrival.com}
\endgroup


\begin{abstract}
In this work, we propose a framework called Auto-Assembly for automated robotic assembly from design files and demonstrate a practical implementation on modular parts joined by fastening using a robotic cell consisting of two robots. We show the flexibility of the approach by testing it on different input designs. Auto-Assembly consists of several parts: design analysis, assembly sequence generation, bill-of-process (BOP) generation, conversion of the BOP to control code, path planning, simulation, and execution of the control code to assemble parts in the physical environment.
\end{abstract}


\section{Introduction}

Assembly planning is one of the most laborious tasks when releasing a new product for manufacturing. Thus, many algorithms and methods around computer-aided design (CAD) and digital twins of the factories have emerged in recent years that help process engineers to prepare a new design for assembly (Computer-aided Assembly Process Planning techniques \cite{caapp}). An emerging trend of Industry 4.0 \cite{industry4.0} suggests that a digital, highly automated factory should be able to infer the process from the design. In practice, even for an automated factory, assembly planning has to be followed by an offline-programming of all the robots and devices to perform the assembly plan.

Additive manufacturing technology (3D printing \cite{3Dprinting}) at the same time has achieved a much higher rate of process design automation. One can simply load a CAD file into a machine that will yield a part of the desired design. The main question we are trying to address in this paper is "Could a robotic cell or even the whole factory work just as a 3D printer?". When loaded with target assembly CAD design and given input parts in specified conditions (e.g. placed in special input jigs) - would it perform the required assembly?

In this work, we show how this can be achieved under specific constraints, paving the road for future experiments towards a more general approach and wider applications. However, the framework we propose is general enough to accommodate more complex designs and conditions, like many types of tooling and different joining technologies.

\begin{figure}[ht]
\centering
\includegraphics[width=0.45\textwidth]{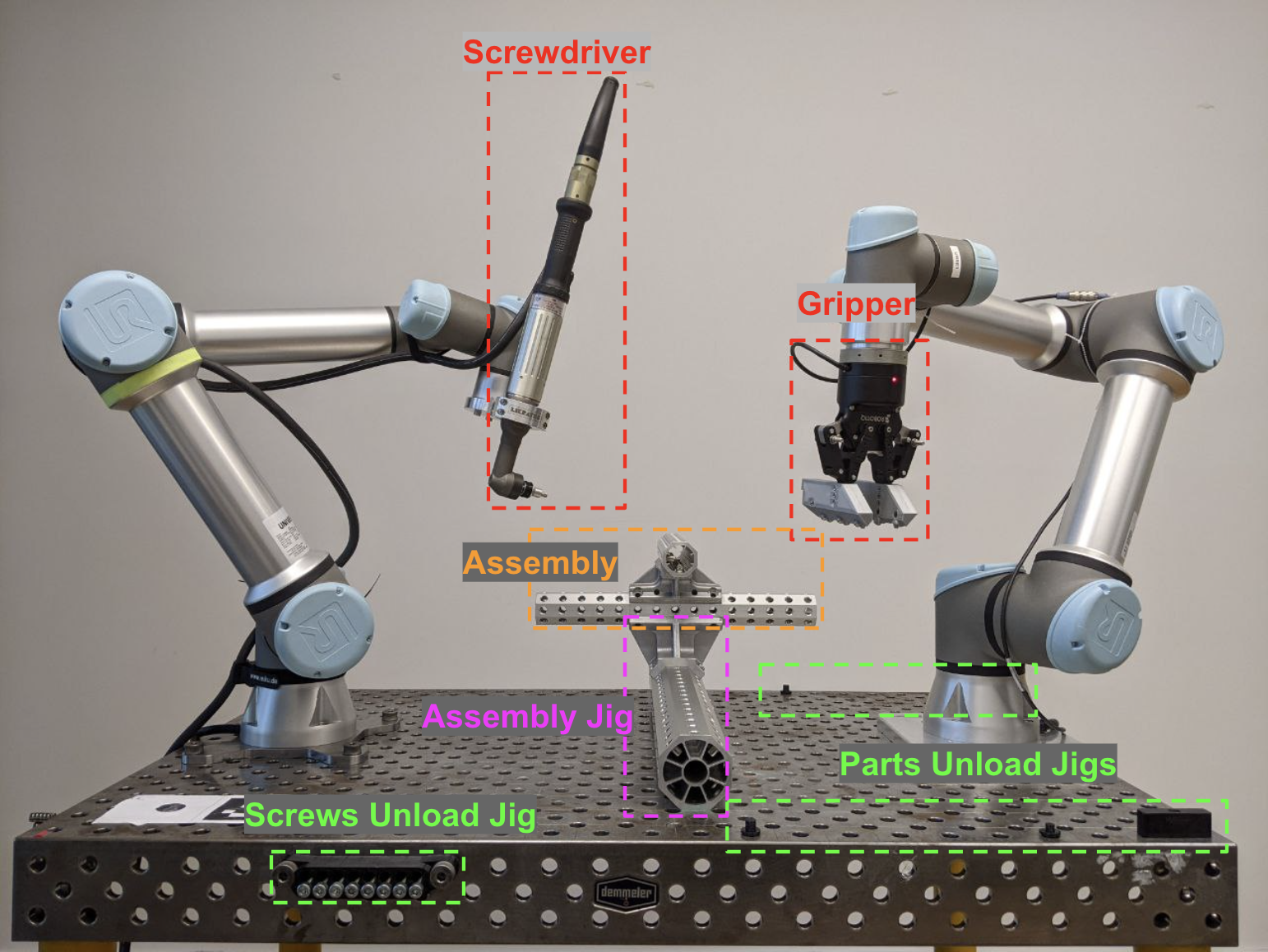}
\caption{Experimental setup: robotic cell with two UR5e manipulators. \textit{Left}: UR5e with a screwdriver Likratec EH2 R1030-A and \textit{Right}: UR5e with gripper Robotiq 2F-85 with custom designed gripper clamps. \textit{On the table}: custom-designed 3D-printed jigs.}
\label{robotic_cell}
\end{figure}
\section{Related Work and Background}  \label{related_work}
An Assembly Planning for a given design typically starts from identifying the mating features or joints and suggesting a feasible Assembly Sequence, which could be automated as seen in \cite{AS_survey}, \cite{ASfromCAD}, \cite{assembly_feature_pair}.

To proceed to the process planning, a virtual environment, also known as Digital Twin \cite{DigitalTwin} is necessary. There are attempts to develop a common ontology, e.g. \cite{ontology}, \cite{mason} and unify interfaces between systems \cite{compas_fab} to support process design automation.

Sierla, Seppo, et al. \cite{AutoML} discuss the conceptual framework of automated assembly planning using a digital twin. It uses the XML-based AutomationML \cite{AutoMLbook} data modeling framework. This framework aggregates different data exchange formats like CAEX for plant description, COLLADA for geometry and kinematics of 3D models, etc.

There is still not sufficient work in joining together process planning, motion planning and execution using a common framework. In \cite{towards_disassembly} authors used artificial intelligence to solve a Tooling Matching problem and developed an add-on for Octopuz\cite{Octopuz} to do a Motion Planning and Robot Program Generation for disassembly, but not testing in physical cells. In another work, \cite{automated_architecture} a similar pipeline is described for an architectural domain, mainly focusing on parametric design and modular assembly.

We claim that Auto-Assembly is the first proposed framework that can generate and execute robotic assembly process for an arbitrary input CAD design.

\section{Problem Statement and Method Overview}  \label{problem_statement}

The main objective of our work is to create a framework that enables a closed loop between design and robotic manufacturing. A target framework should analyse the design and provide a simulation of assembly, executable programs (when possible) and other feedback. The primary aim of the feedback is to help in adapting the design and manufacturing to better correspond to each other.

The feedback we should provide can be split into two categories:
\begin{itemize}
\item Successful simulation and its' artefacts can be directly used to decide on physical manufacturing. Users can choose between different processes to choose the one, based on the key performance indicators (KPI) they want to optimize: time, tooling price, energy consumption, etc.
\item In case of a failure, the system should provide all necessary feedback that helps to change the design, robot's position, choose the robots with better parameters or different cell configuration. Such feedback can be: failed operations, missing appropriate tooling, parts or tools in collision, unreachable states.
\end{itemize}

To achieve this, we implement a framework described in detail in Section \ref{desc}. Section \ref{architecture} gives an overview of our system and its components. Section \ref{input_data} discusses 3D modelling of the assembly design files that form the base of our data extraction pipeline. Section \ref{asg} reviews the usage of this extracted data to produce a set of possible assembly sequences.

Each operation in the assembly sequence is enriched with tooling information as discussed in section \ref{tool_matching}. Finding a specific cell that contains all the resources like jigs, robots, and their tooling, etc to execute all the operations needed for an assembly is explained in section \ref{cell_matching}. Section \ref{code_gen} tells about the generation of the control code that moves the robots to grasp, place and fasten parts in a cell. 

In section \ref{experiments}, we test our framework on different assemblies and discuss the results. In section \ref{conclusion}, we review our findings from the experiments and future work.

\section{Framework}\label{desc}
\subsection{System architecture}\label{architecture}
\begin{figure*}
  \centering
  \includegraphics[width=\textwidth]{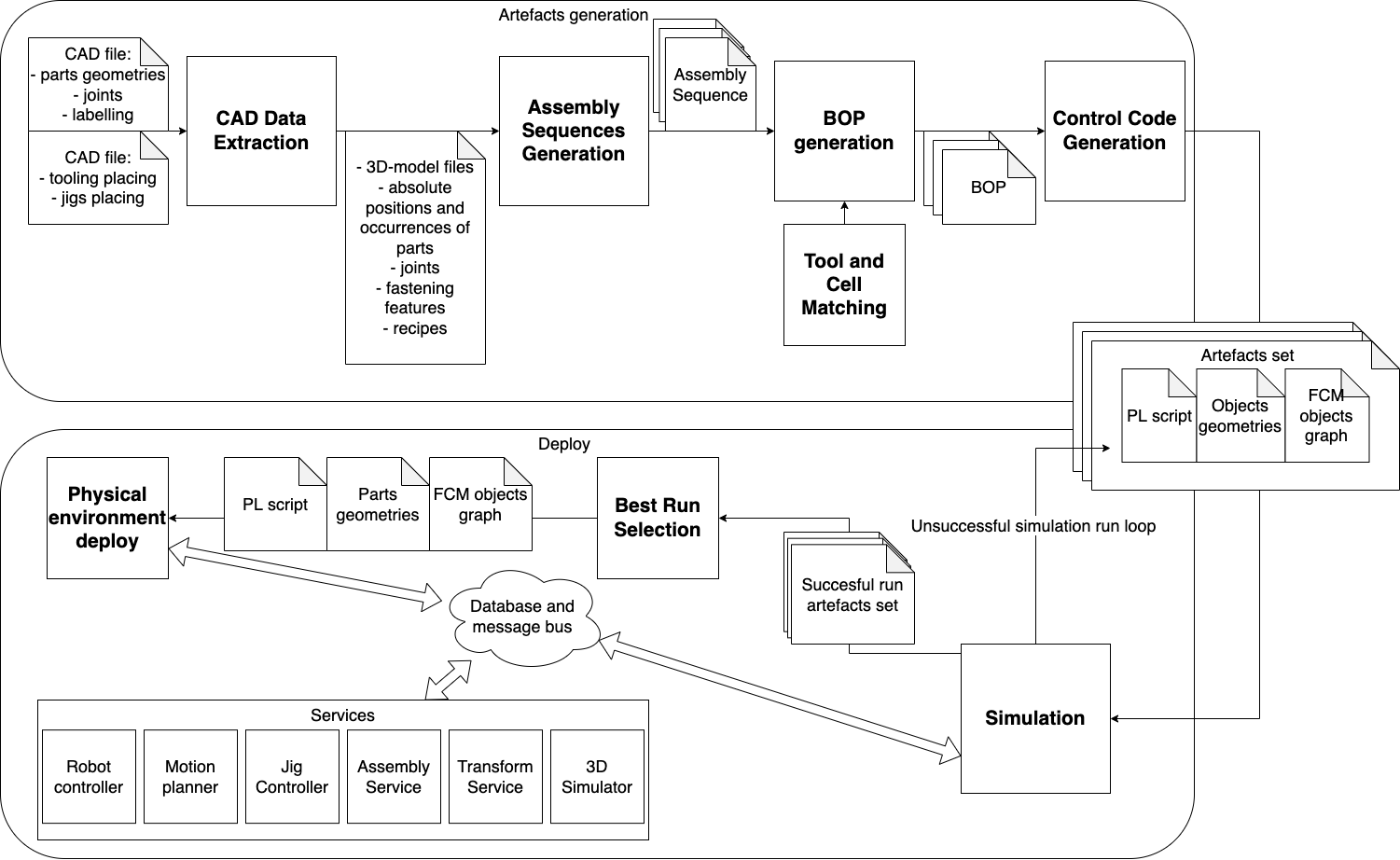}
  \caption{System architecture}
  \label{framework_arch}
\end{figure*}
Auto-Assembly framework can be divided into two parts as shown in Fig. \ref{framework_arch}. The first part, called "Artefacts generation", works with CAD files provided by a design engineer. It is intended to run once on the input data and provide artefacts, which can then be stored and re-used to run the assembly process in the simulated and physical environments. This part includes Assembly Sequence Generation, Tool and Cell Matching, Bill-of-Process (BOP) Generation and Control Code generation.

\begin{figure}[ht]
\centering
\includegraphics[width=0.45\textwidth]{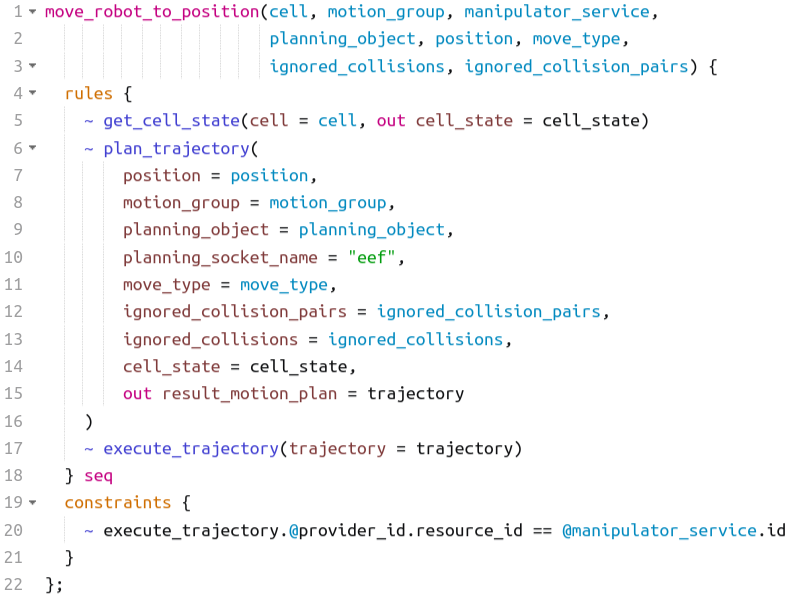}
\caption{Listing of PL-code implementing high-level of robot control. Abilities
\textit{get\_cell\_state} and \textit{plan\_trajectory} are implemented by Motion Planner and \textit{execute\_trajectory} by Robot Controller}
\label{apl_move_robot}
\end{figure}

The second part can be seen as a deployed environment. It is represented as a system where we have many services, providing “abilities” which can be called from the domain-specific Process language (PL). An example of the PL script can be seen in a Listing \ref{apl_move_robot}. Here we describe the most important services and their respective abilities:

\begin{itemize}
\item Robot Controller
    \begin{itemize}
    \item Abilities to control the robots on a low level. As input, it takes a trajectory as a list of a robot’s joint states, and as output, interpolates the trajectory and moves the robot.
    \item Abilities to control tooling connected to the robot, like grippers, screwdrivers, etc.
    \end{itemize}
\item Motion Planner
    \begin{itemize}
    \item Ability to plan a trajectory in the cell to move a robot to a target pose with cell objects taken as the collisions.
    \end{itemize}
\item Jig Controller
    \begin{itemize}
    \item Ability to return a pose of a part in a jig with respect to the jig origin.
    \end{itemize}
\item Assembly Service
    \begin{itemize}
    \item Ability to retrieve the information about fasteners and resulting parts' pose with respect to the cell origin.
    \end{itemize}
\item Transform Service
    \begin{itemize}
        \item Ability to get the position of any object inside a cell with respect to any object in the cell.
    \end{itemize}
\item 3D Simulator
    \begin{itemize}
        \item Abilities to load objects from cell description and visualize cell state.
    \end{itemize}
\item Database and Message Bus
    \begin{itemize}
        \item Abilities to publish and retrieve JSON objects. This component is used as a message bus and data storage.
    \end{itemize}
\end{itemize}

All system parts exchange the data in a special format called Factory Control Model (FCM). It can be considered as a schema and also is a vital part of our system since it lets all the components speak the same language.

\subsection{CAD Data Preparation and Extraction} \label{input_data}

For any given assembly, our framework needs two design files. 
\begin{itemize}
    \item Design file containing part assembly with joints. Fasteners are labelled as separate joints in order to distinguish them from other parts.
    \item Design file containing the jigs and gripper at different stages of assembly like grasping, placing, etc.
\end{itemize}

The examples of these files for an assembly are depicted in Figs \ref{assembly1} and \ref{recipe} and are created by us in Fusion 360. Our method is CAD-software agnostic as long as we can extract the CAD data using an API.

From the design file in Fig. \ref{assembly1}, we extract the joints and part occurrences information using Fusion 360 API \cite{Fusion360}. Using this data, a joint register is created that maps every joint to its parts. The joint register follows the FCM schema. 

From the design file in Fig. \ref{recipe}, we extract the pose of gripper occurrence relative to the part during grasping it from the jig and placing it at the assembly state using the Fusion 360 API. We call this data as \textbf{recipes}.

\subsection{Assembly Sequence Generation}\label{asg}

A CAD design contains a lot of important information about the part's geometries, relations, and absolute poses. But what it lacks - the right assembling order - is the key information to move towards the assembled product.
Assembly sequence encodes the order of operations needed to be performed on parts by the robotic cell.
Although the operations can be executed sequentially, assembly sequences are represented by polytree (directed acyclic graph whose underlying undirected graph is a tree).
Not any such tree represents a valid and feasible assembly sequence:
\begin{itemize}
\item only directly joined parts should be neighbours;
\item the order of operations should take into account the geometrical limitations;
\item the number of generated assembly sequences should be reasonably limited. Naturally, it grows exponentially with the number of parts involved. This makes it hard to check all the generated sequences to pick the best one according to some criteria.
\end{itemize}
The assembly sequence generation step aims to solve all three aforementioned issues, providing a limited number of valid sequences. The whole process can be divided into three steps:
\begin{itemize}
\item a liaison graph generation; 
\item assembly sequences generation based on the obtained liaison graph;
\item geometry feasibility checking based on parts geometries
\end{itemize}
This approach we used is described in \cite{ASfromCAD}. Further, the high-level steps, important implementation details, and differences with the original paper are described.
\subsubsection{Liaison graph generation}
The CAD file consists of the individual parts combined together with joints and fasteners. The information about the joints is crucial to accurately determine parts connectivity. Considering the parts as liaison graph nodes, connectivity information transfers into edges in this graph.
We extracted the information about joints and fasteners from the design in CAD software to build up a liaison graph to further analyze it and generate assembly sequences.
\subsubsection{Sequences generation}
An assembly sequence determines the order of operations on parts. The liaison graph itself, being undirected, doesn't set the order of operations in general. But the order should be based on the liaison graph since the latter contains the information about the connectivity in the resulting assembly.
Usually, there are many sequences of operations. \cite{ASfromCAD} describes the approach of extracting all possible assembly sequences from the liaison graph. We followed the suggested approach.
\subsubsection{Geometry feasibility checking}\label{geometrical_feasibility_check}

The geometrical feasibility of an assembly process is the fundamental property, which should be checked first to eliminate irrelevant sequences. These irrelevant sequences could contain, for example, one part to be joined with another part, which is trapped already inside the sub-assembly.
To prevent this, geometrical analysis of sub-assemblies is used. One sub-assembly is translated step-by-step w.r.t another sub-assembly in one of chosen directions until the bounding boxes of the sub-assemblies still intersect and the solid bodies' intersection is checked. If the intersection represents a volume, it's impossible to join the sub-assemblies in the chosen direction, and the remaining directions should be checked.

Choosing the directions of translations alongside step size is important for the result. Due to the nature of assembly parts and their orientation alignment, directions along the main coordinate axes work well in the tested assemblies. In other cases, information from joints from the CAD file could be used to determine the potential directions.
Step size is computed based on the minimal size of the part across both sub-assemblies. Precisely, the step size is computed as a 0.75 ratio of the diagonal of the smallest bounding box part. The idea behind this value is to exclude the possibility of going completely through the smallest part with a single translation step.

\subsection{Tooling Matching}\label{tool_matching}
The assembly sequence in itself doesn’t require specific tooling models, but this is information is required for the next steps in the assembling process. Given a graph of the assembly sequence from the previous section, we traverse this graph, considering the type of operation and parts used, assigning all the tooling models and adding recipes to process this operation.

To archive this, we extract the following information from the CAD files:
\begin{itemize}
\item For grippers:
\begin{itemize}
\item Model of the part gripper can be applied.
\item List of positions for grasping the part, calculated with respect to the part origin. We use the information from “joints”, such as JointAxis, to extract the vector of connection. Based on this vector poses are calculated.
\item States of digital inputs register to control the gripper.
\end{itemize}
\item For jigs:
\begin{itemize}
\item Model of the part jig can hold.
\item Position of a part in a jig.
\end{itemize}
\item For screwdrivers:
\begin{itemize}
\item Screw-picking requirements, such as type of screw-holder.
\end{itemize}
\end{itemize}

We store this data in Tooling Database. In our approach Tooling Database is a storage with an API which allows adding, matching and visualizing of the tooling.

By analyzing the dataset of the tooling used in physical world production in the automotive field, we concluded that the same information is stored in the tooling design files and propagated to the tooling integration in the physical cells. We decided to formalize the requirements and then store this data. For the cases where it can’t be calculated from the design files, we can manually put that information into the tooling database.

\subsection{Cell matching}\label{cell_matching}
Cell description includes all the information representing an assembly cell. Cell description is used to deploy both environments (virtual and physical) and to choose a cell to execute Assembly Sequence. We topologically sort a graph of the Assembly Sequence and assign a level for every operation. The level is required to assign resources for the parallel operations when we should use different resources of the same model. Then by traversing each operation, we check the resources’ models required for this operation and find their representation in the cell. If there are no cells satisfying all the resource requirements for the Assembly Sequence, we fail, providing feedback with the exact operation and the resources model we were not able to assign. As a result of the execution of the described algorithm, we have an assembly sequence to be converted into a BOP.

\subsection{Control code generation}\label{code_gen}

\begin{figure}[ht]
\centering
\includegraphics[width=0.45\textwidth]{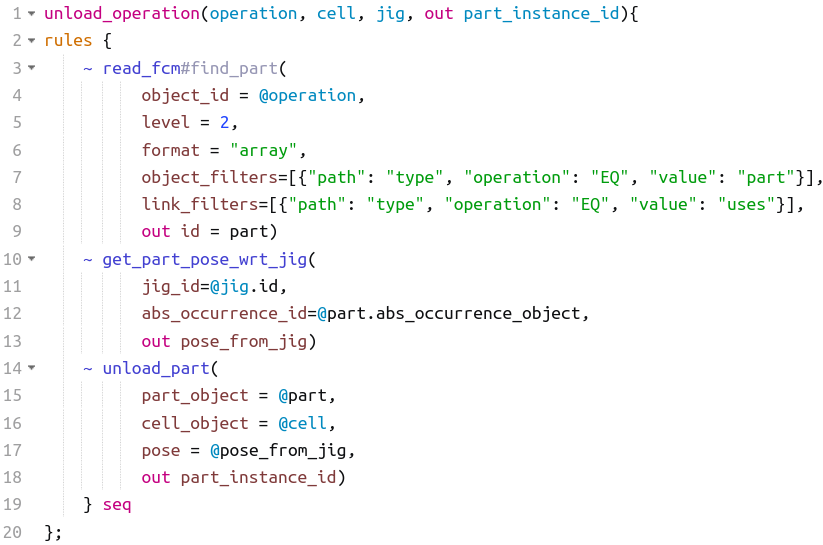}
\caption{Listing of PL-code for unload operation}
\label{apl_unload}
\end{figure}

For each operation in the BOP, we match a specific PL-script, which is self-containing to perform this type of operation, and pass
the operation, its resources and parts as the parameters, creating one PL-script, to assemble a product.
An example of PL-script implementing unload operation is presented on the Listing \ref{apl_unload}

\section{Experiments and Results} \label{experiments}

The objectives of our experiments are:
\begin{itemize}

\item To evaluate the framework. 
\item To evaluate the assembly BOPs in the physical environment to provide metrics and feedback on the assembly. 
\end{itemize}

\begin{figure}[ht]
\centering
\includegraphics[width=0.3\textwidth]{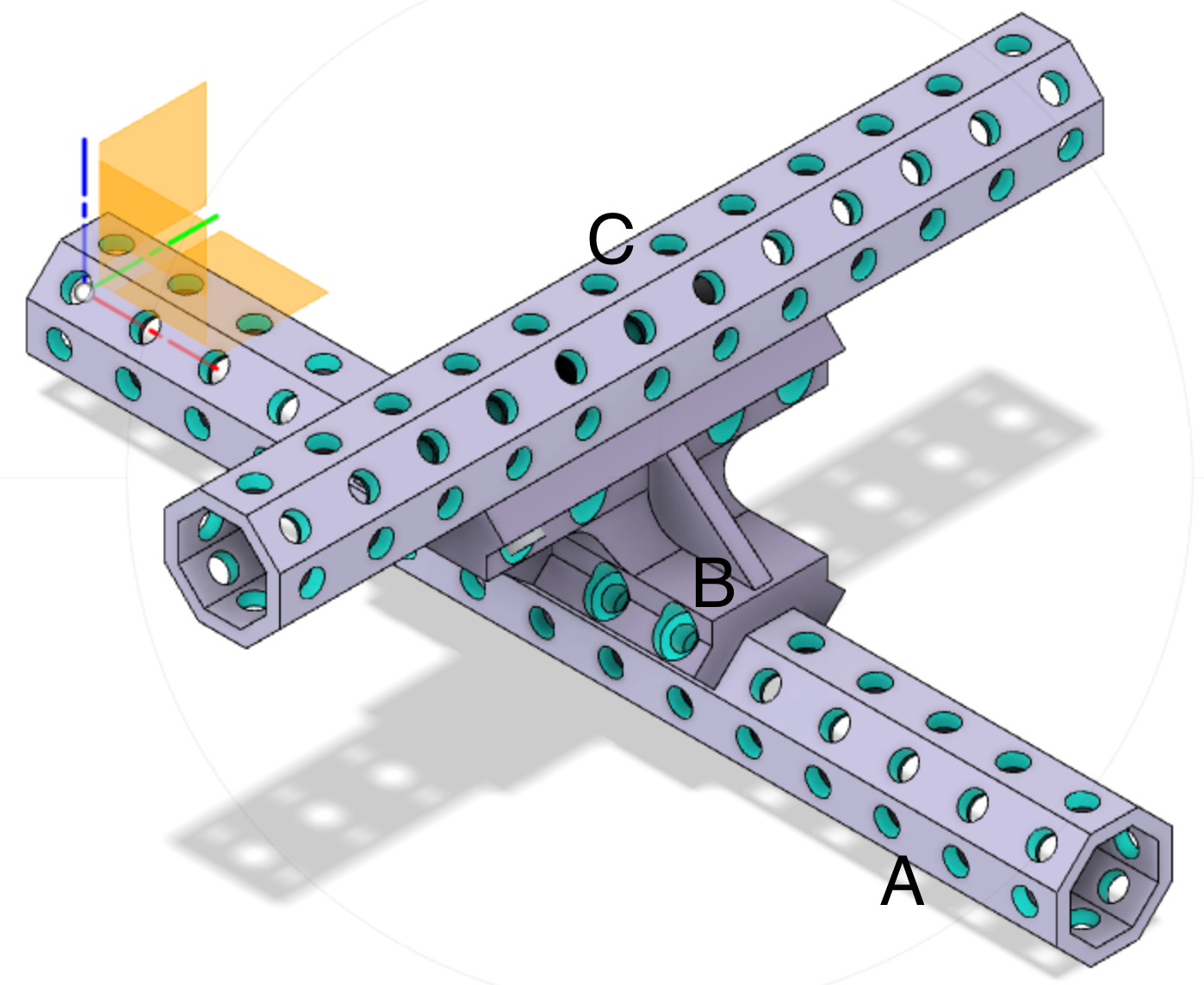}
\caption{A simple assembly containing 3 parts. \textit{Profiles}: A, C and \textit{connector}: B}
\label{assembly1}
\end{figure}

\begin{figure}[ht]
\centering
\includegraphics[width=0.3\textwidth]{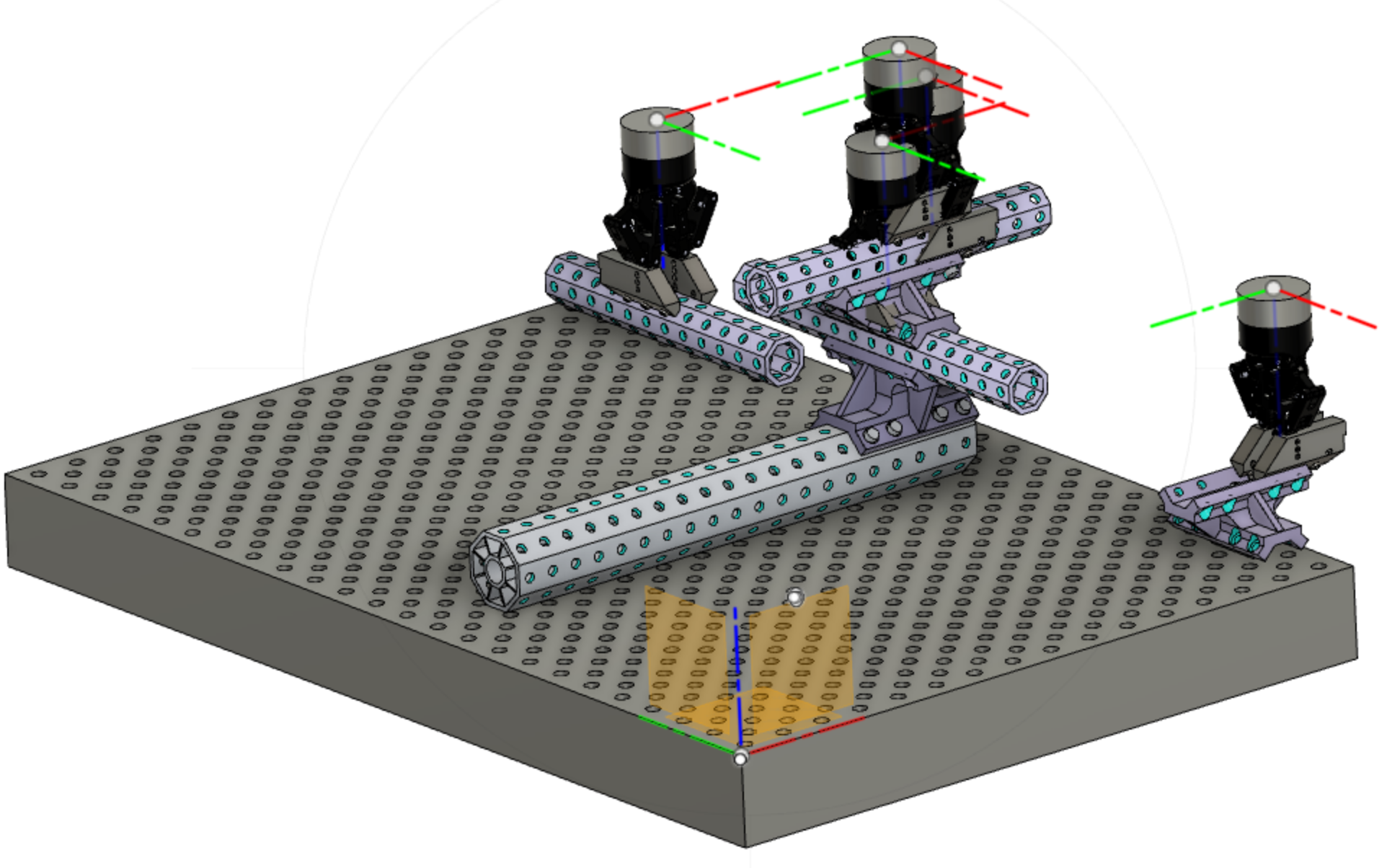}
\caption{Assembly Design file showing the assembly jig, custom-designed gripper adapters, grasping, and insertion states of the gripper. \textit{Assembly state}: Center of the table. \textit{Jig state}: Top and bottom right of the table.}
\label{recipe}
\end{figure}

The assembly we chose to test is shown in Fig. \ref{assembly1} and its tooling, and jig design are shown in Fig.\ref{recipe}.

\begin{itemize}
\item \textbf{Data Preparation}:
\begin{itemize}
\item we extract the joint register and recipes as mentioned in section \ref{input_data}.
\item taking the joint register and part models files, assembly sequence generator produced 8 assembly sequences for this assembly. These sequences are mentioned in Fig. \ref{asg_diag}.
\item we enrich the assembly sequences using tool matching mentioned in section \ref{tool_matching}.
\item we convert the enriched assembly sequences to BOP using cell matching as mentioned in section \ref{cell_matching}.

\end{itemize}

\begin{figure}[ht]
\centering
\includegraphics[width=0.4\textwidth]{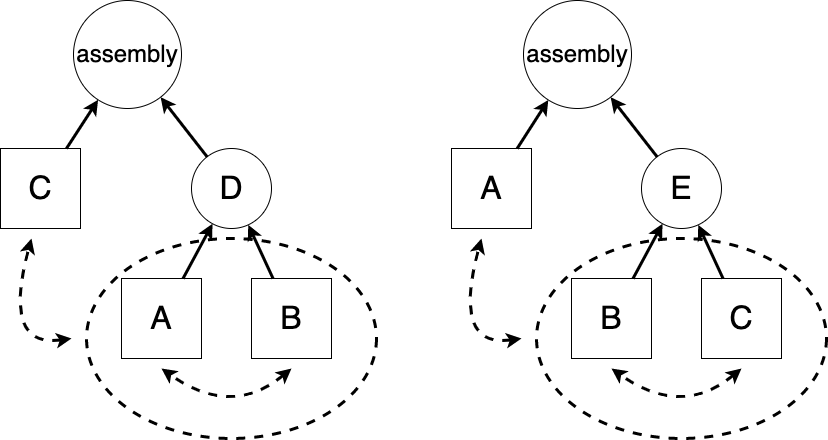}
\caption{A diagram of assembly sequences generation process for the design used in the experiments. Square blocks represent parts while circles represent (sub-)assemblies(D and E). The possible sequences are \textit{Left}: ABDC, BADC, CABD, CBAD and \textit{Right}: BCEA, CBEA, ABCE, ACBE.}

\label{asg_diag}
\end{figure}

\item \textbf{Scene Preparation}: Before starting the assembly, if it's a simulated environment, the jigs are unloaded at the same poses as in the physical world in Fig. \ref{robotic_cell}. If it's the physical environment, the jigs and parts are placed in their respective poses.  

\end{itemize}

\begin{figure}[ht]
\centering
\includegraphics[width=0.45\textwidth]{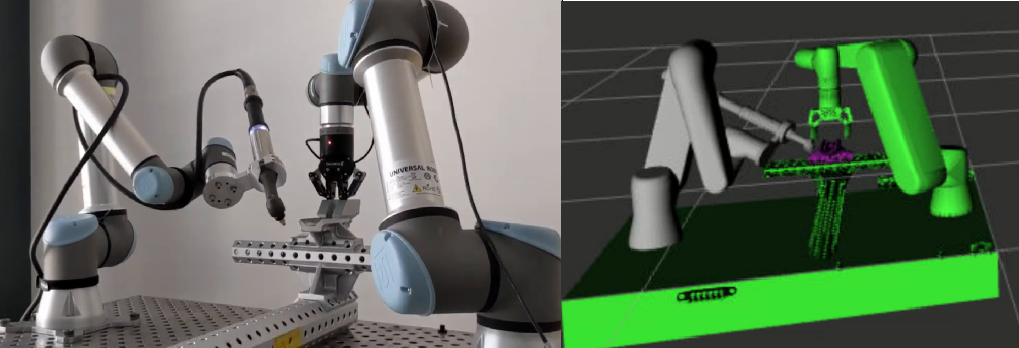}
\caption{Assembly process. \textit{Left:} Physical environment and \textit{Right:} Virtual environment.}
\label{virtual-and-real-assembly}
\end{figure}

\begin{itemize}
\item \textbf{Simulation}

\begin{enumerate}

\item Start simulation deployment with services and a database and message bus instance as mentioned in Fig. \ref{framework_arch}.
\item We trigger execution of the PL code, which starts from running operations of type "unload" on input parts. This operation effectively initializes part instances on respective positions in the input jigs, so that the parts are now represented in the digital twin of the cell, as active objects with poses, visible for the simulator as well as for the motion planner.
\item The rest of the PL code is executed, sequentially reading necessary gripper positions, planning and executing trajectories, and triggering gripper control programs for grasping/releasing/fastening, all by calling respective PL functions that use abilities of underlying systems described in \ref{architecture}.
\item The user can observe the execution of assembly in the 3D simulator. During the execution of the assembly process, the motion planner gives us direct feedback, on whether it can reach a certain pose in the assembly or not.

\end{enumerate}

 For our example assembly, the initial results were the following:
\begin{itemize}
    \item Of all the possible 8 assembly sequences, only one assembly sequence (ABDC) passed through the cell matching, as the cell resource descriptions (in this case jigs) support this. Many sequences get filtered based on the cell resources (jigs, robot tooling, etc.). 

    All the sequences for any given assembly are feasible, if the cell has the resources to hold sub-assemblies, For example, the assembly sequence CABD is possible when the cell has the jig that supports moving part C first to the assembly pose, then creating a sub-assembly D by moving parts A and B in the same order.  Now the question here is, how to make the decision on which resources in this case jigs are needed to be designed to hold the sub-assemblies. 
    If there are cycles in the graph, multiple BOPs pass through the cell matching which needs the same cell resources, which enables us to simulate and select the best one based on the metrics.   

           \item We also noticed that our initial design failed due to fastening robot reachability, we took this feedback from the framework and changed the fastening position in the assembly to assemble a product.

\end{itemize}
To adjust the design, we changed the positions of the screws in the assembly to other holes without losing the structure stability. After this design adjustment, we were able to successfully simulate the one feasible assembly sequence.

This is one of the main features of the proposed framework - to get this kind of feedback about the product/tooling/cell design compatibility as soon as possible with minimal manual input.

\item \textbf{Running assembly on physical robotic cell:}
Once we find an assembly sequence that passes in the simulation, we can proceed to the physical assembly process.

This is achieved by running the same generated PL code as before but now in a physical robotic environment. The only thing that differs compared to the previous pipeline in simulation is the first step - deployment of the systems. For a physical assembly we deploy the robot and the gripper drivers to be connected to robots and devices, such that in parallel with updating the state of the digital twin, these controllers will be changing the states of the tools in the physical world, such as robots moving along precomputed trajectories, gripper opening/closing and the screwdriver fastening the screws.

We evaluated this assumption on the physical robotic cell with two collaborative robots the layout of which can be seen in Fig. \ref{robotic_cell}. The results of these experiments are two folds:
\begin{enumerate}
    \item On one side, we can see that as soon as the digital twin is accurate enough, all computed gripper positions allow performing most of the operations, such as picking a screw, grasping and releasing a part, and in some cases to fasten a screw.
    \item On another side, some operations show that an accumulated tolerance stack of robot calibration, tool accuracy, and parts accuracy leads to the inability to perform the joint operation such as fastening successfully, and the screwing position requires correction.
\end{enumerate}
The example of running assembly in the virtual and physical environments can be seen in Fig. \ref{virtual-and-real-assembly}.
The process of assembly of the provided CAD by running the generated PL code can be seen in the accompanying video.

\end{itemize}

\section{Conclusion and Future Scope}\label{conclusion}

In this paper, we implemented and tested a framework to run a robotic assembly of a product by using only CAD files as input. We were able to use the feedback provided by the framework to change the original design and achieve successful assembly. We re-iterated the whole pipeline and transferred the assembly from the virtual to the physical world. We conclude that this transfer can be done only if the digital twin matches the physical cell precisely, which requires additional work, such as robots and cell calibration, but it's out of the scope of this paper. The choice of design of our system proved its flexibility since we were able to analyze and change artefacts produced during the different steps of the execution. The system is general enough to support new products and cell configurations.

The next step is to validate our framework on more complex assemblies, including new types of operations and operations which involve more than two parts.

In our experiment, we relied only on the parts' dimensional precision and the accuracy of the robots. While it could work for some parts, and partially worked in our case, it would likely fail on many other parts and materials. To address this problem, computer vision and other perception methods should be introduced into the framework to deal with variations in the real assembly process.

In the section \ref{geometrical_feasibility_check} the constraint we chose could lead to some possible assembly sequences being rejected. To solve this issue, we plan to implement a geometrical feasibility check based on joints from the CAD files or other optimization algorithms.

The method used in the section \ref{cell_matching} can lead to a sub-optimal configuration or even to setups where some robots can't reach parts. This approach was chosen as the easiest to track and implement. In future works, we plan to implement a more sophisticated scheduling algorithm based on geometrical and utilization constraints.


\end{document}